\documentclass[letterpaper, 10 pt, conference]{ieeeconf}  

\usepackage{algorithm}
\usepackage{algorithmic}

\usepackage{amssymb}		
\usepackage{graphicx}		
\usepackage{hypernat}		
\usepackage{amsmath}
\usepackage{comment}
\usepackage{booktabs}
\usepackage{multirow}
\usepackage{bm}
\usepackage[backend=biber,style=ieee,natbib=true]{biblatex}  %
\usepackage[normalem]{ulem}
\usepackage{color}
\usepackage{pgfgantt}
\usepackage{caption}
\usepackage{rotating}
\usepackage{subfig}
\usepackage[font=small,labelfont=bf]{caption}
\IEEEoverridecommandlockouts                              

\usepackage{graphics} 
\usepackage{epsfig} 
\usepackage{mathptmx} 
\usepackage{times} 
\usepackage{amsmath} 
\usepackage{amssymb}  

\title{\LARGE \bf
RMap: Millimeter-Wave Radar Mapping Through Volumetric Upsampling}

\author{Ajay Narasimha Mopidevi, Kyle Harlow, and Christoffer Heckman$^{1}$
\thanks{$^{1}$All authors are with the Department of Computer Science,
        University of Colorado Boulder, Boulder CO, USA
        {\tt\small christoffer.heckman@colorado.edu}}%
}

\begin{document}

\maketitle
\thispagestyle{empty}
\pagestyle{empty}

\begin{abstract}
Millimeter Wave Radar is being adopted as a viable alternative to lidar and radar in adverse visually degraded conditions, such as the presence of fog and dust. However, this sensor modality suffers from severe sparsity and noise under nominal conditions, which makes it difficult to use in precise applications such as mapping. This work presents a novel solution to generate accurate 3D maps from sparse radar point clouds. RMap uses a custom generative transformer architecture, UpPoinTr, which upsamples, denoises, and fills the incomplete radar maps to resemble lidar maps. We test this method on the ColoRadar dataset to demonstrate its efficacy. 

\end{abstract}

\section{Introduction}
\label{sec:Introduction}
Many recent advancements in navigation use lidar or camera sensors for odometry estimation and mapping \cite{ebadi2022present}. While these sensors provide highly reliable data under normal conditions, they are severely affected in visually degraded environments (VDEs) such as fog, smoke, rain, and snow. In VDEs, lidars can provide false returns, requiring heavy filtering for mitigation \cite{burnett2022we}, and in cameras, the geometry used to track features can be obscured \cite{kramer2021radar}. 

\begin{figure}[!t]
    \centering
    \includegraphics[width=0.48\textwidth]{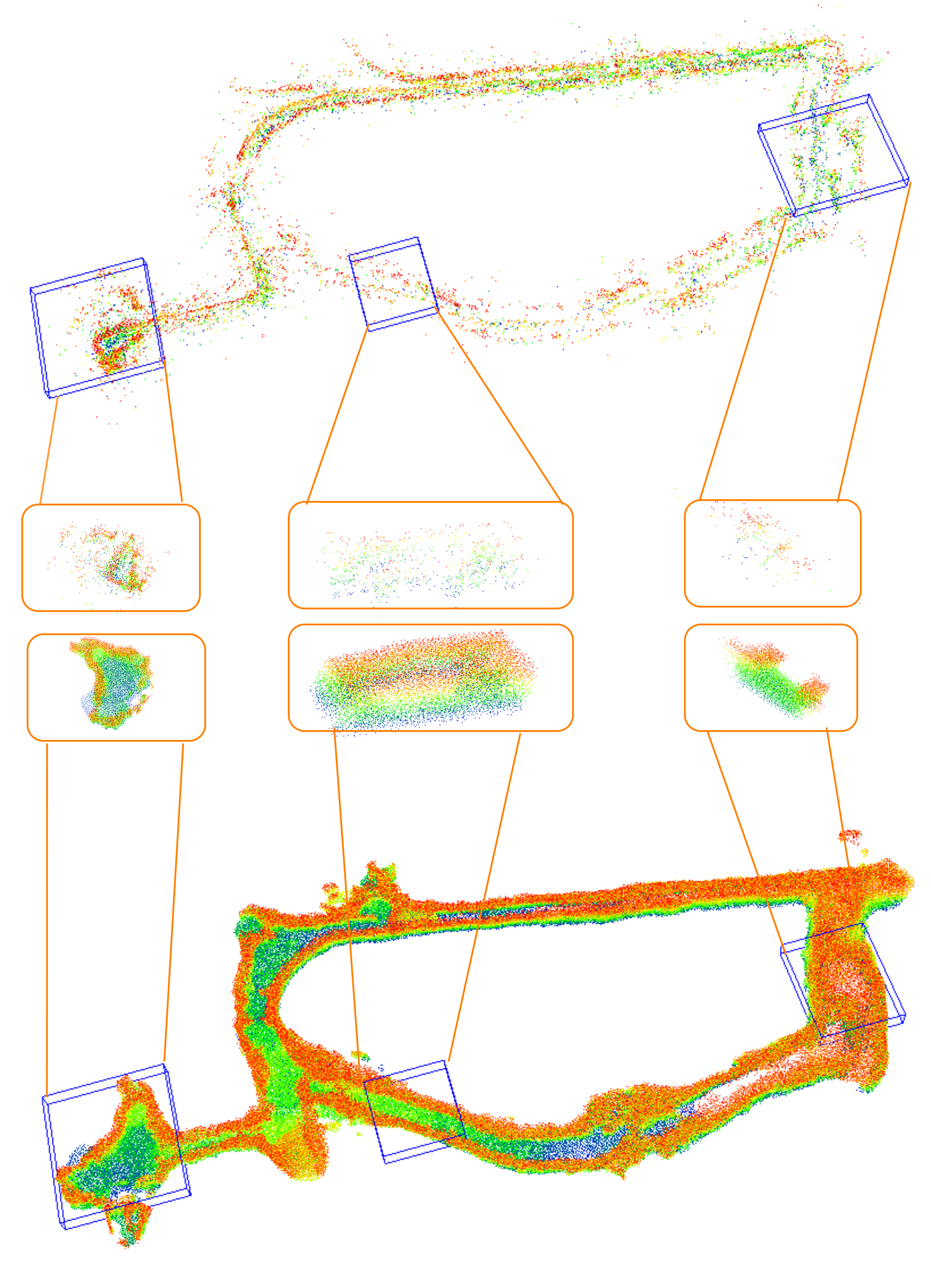}
    \caption{A qualitative comparison between a map generated using just the Radar Octomap method \cite{kramer2021radar}, and RMap, our novel upsampling method. The top row represents the initial radar map generated with Radar Octomap, and the bottom row represents the map generated using RMap. The patches show our method filtering noise from regions of free space, as well as upsampling and filling sparse and empty regions respectively.}
    \label{fig:results}
    \vspace{-7mm}
\end{figure}

Millimeter-wave (mmWave) radar sensors are a compelling alternative in adversarial conditions as their longer wave lengths allow measurements to bypass particulate clutter \cite{lu2020see}. Common automotive-grade mmWave radar operates between 76-81 GHz and can provide centimeter-level accuracy signal detections. 


However, mmWave radar suffers from several key drawbacks compared to other ranging sensors such as lidar. Radar is susceptible to various types of noise, resulting from constructive and destructive interference from the projected wave patterns reflecting off the environment. These wave patterns can lead to spurious returns, lost returns, and multi-path reflections. Additionally, radar data is often sparse, especially when examining the constant false alarm, non-maximal suppression filtered targets. For example, automotive-grade imaging radar typically produces a maximum of just a few hundred points per scan. Furthermore, radar systems generally have significantly lower resolution compared to competing lidar systems.

Traditionally, radar applications have been limited to collision avoidance for autonomous vehicles, rather than high-precision tasks such as odometry estimation and mapping. However, recent research has led to the development of robust systems capable of generating odometry and filtering radar data for mapping purposes. We present one such solution, RMap (RadarMapping), a method to generate high-precision 3D maps using radar point clouds extracted from an mmWave sensor. We present an end-to-end pipeline for generating the 3D maps from radar point clouds and demonstrate how these maps can be leveraged to construct a 3D map resembling lidar-based maps through deep learning techniques. We show examples comparing the original radar maps, lidar maps, and the map generated on radar data using our method in Figure \ref{fig:results}.

Our contributions to the state-of-the-art include:

\begin{itemize}
    \item We have developed a Pose-based region sampling algorithm, which effectively represents the 3D maps as partial point clouds.
    \item We introduce UpPoinTr, a generative transformer architecture, that performs upsampling, denoising, and fills the incomplete maps.
    \item Utilizing the radar sensor model for octomap \cite{hornung2013octomap, kramer2021radar}, we have generated 3D radar maps for the ColoRadar dataset and processed patches of these maps through UpPoinTr to closely resemble lidar-based maps after reassembly.
    \item Our work represents the first known application of point cloud completion networks to radar-based 3D scene generation tasks. 
\end{itemize}


\begin{figure*}
\vspace{2mm}
    \centering
    \includegraphics[width=0.98\textwidth]{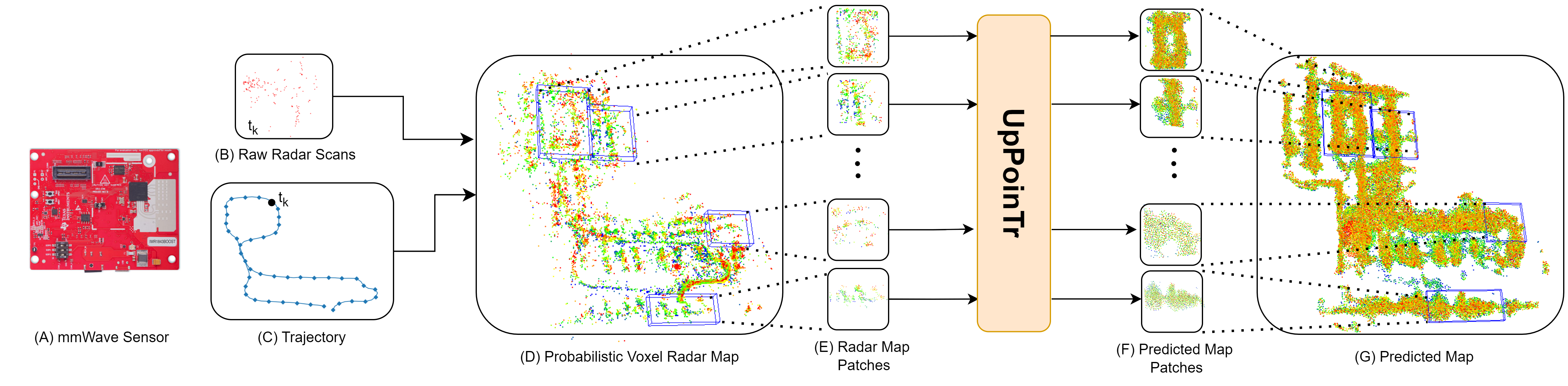}
    \caption{System Diagram of RMap. An (A) automotive-grade system-on-chip mmWave radar sensor produces (B) Radar point clouds generated by CFAR thresholding along a trajectory (C). A (D) Radar Map is generated using the radar point clouds with Radar Octomap along the trajectory of the robot, with different pose locations highlighted. This map is then divided into (E) Patches of size 2048 sampled at different locations along the trajectory. These patches are passed through our upsampling network, UpPoinTr, to produce (F) predicted map patches. The predicted point clouds of size 8192 from UpPoinTr are then merged into (G) the final predicted radar map. This method is described in detail in Section \ref{sec:methods}}
    \label{fig:System diagram of RMap}
     \vspace{-7mm}
\end{figure*}

\section{Background}
\label{sec:related_works}

As radar is still a new sensor to robotics we provide a primer on frequency-modulate continuous-wave (FMCW) radar and then analyze related works to both mapping with radar along with current architectures for point completion. 

\subsection{Radar Primer}
FMCW radar operates by transmitting a sequence of chirps, or waves with linearly increasing frequency from a series of antennae. These signals are reflected back from the environment and then captured by another series of receive antennae. A mixer subtracts the transmitted signal from the received signal, generating an intermediate frequency correlated along individual chirps, across chirps, and by each transmit and receive antenna. The series of digital samples from an analog-to-digital converter resulting in these measurements is referred to as a data-cube \cite{kramer2022coloradar}. 

After assembling the data-cube fast-Fourier transforms can be run along each dimension of the data-cube to generate range-measurements, range-rate measurements, and angle-of-arrival measurements from the frequency domain \cite{TI_mmWave_fundamentals,richards2010principles,melvin2013principles}. In most commercially available SoC radar, the resulting frequency graphs are post-processed by a constant false alarm rate filter \cite{hm1968adaptive,melvin2013principles} which uses non-maximal suppression to determine sufficiently unique point targets published into a point cloud for end-use applications. As mentioned in Section \ref{sec:Introduction}, these targets are subject to various sources of noise and are more sparse than other range-measurement sensors. 

\subsection{Related Works}
\textbf{Mapping:}
Approaches to radar-based mapping usually seek to address inherent radar challenges, such as susceptibility to false positives stemming from spurious returns and multipath reflections. \cite{weston2019probably} attempted to create an occupancy grid by learning an inverse sensor model, incorporating modeled variance to account for areas occluded in the sensor's field of view. This work also relied on 2D scanning radar. \cite{lu2020see} used generative adversarial networks trained with lidar scans to generate radar grid-maps, which were robust to fog, and filled in areas from sparse radar measurements. \cite{park2019radar} applied target presence probabilities to radar scans aligning them to map priors to generate 2D grid maps. Another recent method used a U-Net to learn super-resolution range-azimuth measurements for 2D grid-maps \cite{prabhakara2023high}. 

While the majority of these methods have primarily operated in 2D, more recent approaches have aimed to tackle the challenge of creating 3D maps. \cite{xu2022learned} learned depth representations from dense radar heatmaps, leveraging this information to generate 3D radar maps. \cite{zhuang20234d} generated a full SLAM solution with correlated key points extracted from multiple scans used to generate 3D maps. 

Our approach is focused on generating 3D maps using sparse radar points, rather than heatmaps, with common occupancy mapping methods such as Octomap \cite{hornung2013octomap}. This still creates two significant challenges: 1) the beam sensor model used in Octomap does not accurately model the operation of the radar sensor, and 2) radar false positives are still present. To address the first problem, we employ a new, radar-specific sensor model first presented in \cite{kramer2021radar}. To address the second challenge, we leverage point cloud completion networks to upsample, fill, and denoise radar point clouds. 

\textbf{Pointcloud Completion Networks:}
Early point cloud networks such as PointNet, use multi-layer-perceptron layers to aggregate features through farthest point sampling (FPS) and achieve permutation invariance on unstructured point cloud \cite{qi2017pointnet,qi2017pointnet++}. The first proposed point cloud completion network (PCN) uses convolutional encoders to transform the input point cloud to a k-dimensional feature vector, and the decoder utilizes this feature vector to generate both a coarse and detailed point cloud \cite{yuan2018pcn}. 

More recently transformers \cite{vaswani2017attention} have revolutionized many types of networks including point cloud completion networks. In their seminal method, PoinTr\cite{yu2021pointr} extracts a coarse point cloud and features using DGCNN graph extractor and uses these resulting features in a geometry-aware transformer to compute set-to-set translational features for point cloud completion. These translational features are employed to generate coarse-to-fine point cloud upsampling steps using FoldingNet\cite{yang2018foldingnet}. Improving on their method, Yang developed AdaPoinTr which uses the PoinTr backbone and introduces dynamic point query generation to rank the input point proxies and the encoder outputs, and denoised queries to make the decoder learn from high-quality queries \cite{yu2023adapointr}. To mitigate the memory constraints associated with analyzing full pointcloud SeedFormer extracts local coarse point cloud seeds and applies an upsampling transformer to gradually refine coarse-to-fine pointcloud through 3 stages. In UpPoinTr, we combine the base architecture of AdapoinTr with the point cloud region upsampling of SeedFormer as described in Section \ref{sec:methods}


\section{Methods}
\label{sec:methods}
Our method consists of three steps, as illustrated in Figure \ref{fig:System diagram of RMap}. First, a series of radar scans and a trajectory are passed through a probabilistic map generator. The resulting map is then divided into patches using a trajectory-based region sampler. Finally, these patches are passed through our upsampling network, UpPoinTr, and re-assembled into a predicted lidar-like map. We provide further details of these steps in the following subsections. We also provide our code at \url{https://github.com/arpg/RMap}.

\subsection{Probabilistic Voxel Map Generation}\label{sec:voxel}
Since radar scans contain a limited number of points compared to standard 64-beam lidar scans which can contain up to $2^{16}$ points, directly upsampling radar scans to match lidar scans is challenging. Additionally, radar and lidar have different fields of view (FoV). Therefore, our first step is to create an intermediate representation bridging the gap between radar and lidar. Maps serve as this intermediate representation allowing for the accumulation of various scans into the structures of their relative environments. Voxel-based maps allow us to generate a space-efficient representation of the world. We specifically use Octomap \cite{hornung2013octomap} for both lidar and radar-based maps. 

To accurately compare these maps, we first apply a filter to the lidar, restricting the FoV to $180^{\circ}$ to match the FoV of the radar. Subsequently, we generate a lidar octomap with the filtered point cloud and a ground-truth trajectory. We then generate a radar map using the radar sensor model presented in \cite{kramer2021radar}. This sensor model replaces the ray-cast model from octomap with an explicit radar occupancy model. Voxels within the radar FoV where radar targets are detected have their probability increased, while voxels in the FoV without returns have their probability decreased. This model more accurately captures occupied space considering the noisy yet penetrative capabilities of radar sensors \cite{kramer2021radar}. 

We have observed that these radar maps consistently have fewer voxels than lidar maps. Across the ColoRadar dataset, lidar maps typically contained approximately 2x--7x more voxels than the radar map applying only the radar sensor model. To address this disparity, we have designed an upsampling and filling method with the aim of bridging this gap and generating lidar-like maps from radar scans.

\subsection{Pose-based Patch Sampling} 
\begin{algorithm}[htb]
\caption{Patch Sampling Algorithm}\label{alg:sampling}
\begin{algorithmic}
\REQUIRE $\text{Lidar Map: } \mathcal{P}^L = \{p_i | p_i \in \mathcal{R}^3\ \text{for } i=1,...,n^L\}$\\
$\text{Radar Map: } \mathcal{P}^R = \{p_i | p_i \in \mathcal{R}^3\ \text{for } i=1,...,n^R\}$\\
$\text{Trajectory: } \mathcal{X} = \{x_j | x_j \in \mathcal{R}^3\ \text{for } j=1,...,m\}$\\
$\text{Seedpoint Threshold: } p_{Th}$\\ 
$\text{Lidar patch size: } N^L$\\
$\text{Radar patch size: } N^R$\\
$\text{Sub-Patch Size: } aN^L$
\ENSURE $\text{Lidar Patches: } \mathcal{P}_k^L \subseteq \mathcal{P}^L$\\
$\text{Radar Patches: } \mathcal{P}_k^R \subseteq \mathcal{P}^R \text{ for } k=1,...,t*m$
\\
\STATE 
\STATE Initialize Lidar-patches $\mathcal{P}_k^L = \{\}$ , Radar-patches $\mathcal{P}_k^R = \{\}$\\
\STATE Generate seed points $p_{s} \subseteq \mathcal{X}$ such that $|x_i-x_j|>p_{Th} \text{ for } i=1,...,m \text{, for } j=1,...,m \text{ and } i\neq j$

\FOR{$p \in p_{s}$}
    \STATE Set $\mathcal{P}_b$ equal to the closest $aN^L$ points in $\mathcal{P}^L$ for $p$
    \STATE Sample the $t$ furthest point in $\mathcal{P}_b$ to generate anchor points $p_{a,i}$
    \FOR{i = 1:t}
       \STATE Find $N^L$ closest points to $p_{a,i}$ of $\mathcal{P}^L$
       \STATE Append result to $\mathcal{P}_k^L$
       \STATE Find $N^R$ closest points to $p_{a,i}$ of $\mathcal{P}^R$
       \STATE Append result to $\mathcal{P}_k^R$
    \ENDFOR
\ENDFOR
\STATE Return $\mathcal{P}_k^L$ \text{, } $\mathcal{P}_k^R$
\end{algorithmic}
\end{algorithm}
While the size of each radar octomap is dependent on the length and shape of its trajectory, most maps contain more than $2^{16}$ voxels. If each voxel's center point is a single input to the generative transformer, memory limitations on modern graphics cards make inputting the entire map infeasible. To effectively train and run the network at inference, we aim to generate patches containing orders of magnitude fewer points. Randomly generating these patches may leave some areas uncovered. Notably, with random sampling, region-growth algorithms selecting the nearest $n$-points from a given pose sample could miss edges and corners of the map. 

Instead, we present a pose-based patch sampling algorithm. Our algorithm jointly generates patches between the lidar-ground truth maps and radar maps. It can easily be modified to remove the reliance on lidar. As input our algorithm takes both a lidar map, a radar map, and the trajectory along which these maps were generated. The trajectory is sampled to generate, what we call, seed points, which are points along the trajectory separated by a tunable threshold. For each seed point, we generate a patch on the lidar map, sampling the nearest $aN^{L}$ points. From this patch, we generate anchor points by sampling the $t$ farthest points along this patch. Subsequently, for each anchor point, we use geometric region growing to sample either the $N^{L}$ or $N^{R}$ nearest lidar and radar points. These resulting sub-patches are accumulated into the set of lidar patches $\mathcal{P}^L_k$ and radar patches $\mathcal{P}^R_k$. These sets of patches cover a sufficient portion of the map, with all points being captured in at least one patch. We detail the algorithm formally in Algorithm \ref{alg:sampling}.

\subsection{UpPoinTr}
After generating patches, we process them through a custom upsampling point cloud completion network, UpPoinTr. UpPoinTr shares its core architecture with AdaPoinTr \cite{yu2023adapointr}, utilizing the set-to-set translation strategy and the geometry-aware transformer architecture, and dynamic query generation blocks. We enhance the coarse point cloud outputs $\mathcal{P}_{c,0}$ by passing each point cloud through a series of upsample layers inspired by the SeedFormer architecture \cite{zhou2022seedformer}. Each upsampling layer generates a hierarchical series of point clouds, progressively increasing the resolution. Each Upsample layer takes the previous upsample layer's point cloud $\mathcal{P}_{c,i-1}$ and the predicted proxy features $\mathcal{F}_{i-1}$ from the previous layer to generate an upsampled point cloud $\mathcal{P}_{c,i}$ and $\mathcal{F}_{i}$. The first Upsample layer considers the $\mathcal{P}_{c,0}$ as the initial point cloud and the decoder features from AdaPoinTr $\mathcal{F}_0$ as the first feature set.


\subsection{Network Training}
To account for the permutation invariance of the unstructured point clouds, we employ the Chamfer distance (CD) as the loss function\cite{fan2017point}, which has ${O(N \, \text{log}(N))}$ complexity. To capture both the fine details and the overall structure at each upsampling layer, we consider CD loss for each intermediate point cloud generated. The ground truth for each of the intermediate point clouds ($G_i$) is generated by downsampling the ground truth point G to $|\mathcal{P}_{c,i}|$ using FPS. The resulting CD calculation appears as:

\begin{align}
    CD_i &= \frac{1}{|\mathcal{P}_{c,i}|} \sum_{p \in \mathcal{P}_{c,i}} \min_{g \in G_i} \| p - g \| + \frac{1}{|G_i|} \sum_{g \in G_i} \min_{p \in \mathcal{P}_{c,i}} \| p - g \|.
\end{align}

\noindent The overall loss $J$ is:

\begin{align}
    J &= \sum_{i}CD_i.
\end{align}

Using this loss function, we trained UpPoinTr on the ColoRadar dataset \cite{kramer2022coloradar} which contains a suite of radar and lidar point clouds on a diverse set of 44 scenes captured across 6 different indoor and outdoor environments. We generated voxel maps as described in Section \ref{sec:voxel} using a 0.15m resolution, a typical voxel size used in navigation applications \cite{ohradzansky2021multi,biggie2023flexible}. We then converted the maps into point clouds using the center of each occupied voxel in the map. We detail the number of scenes, along with the ratio of lidar map points to radar map points (L/R) for each environment in Table \ref{tab:dataset}. 

\begin{table}
    \vspace{2mm}
    \begin{tabular}{|p{2cm}||p{0.75cm}|p{0.75cm}|p{3.2cm}|}
        \hline
        \textbf{Environment} & \textbf{\# Scenes} & \textbf{L/R Ratio} & \textbf{Deviation distribution [50\%, 75\%, 90\%, 95\%] (m)} \\
        \hline \hline
        arpg\_lab & 5 & 4.28 & [0.150, 0.450, 0.808, 1.500] \\
        \hline
        aspen & 12 & 2.59 & [0.260, 0.450, 0.654, 0.862] \\
        \hline
        ec\_hallways & 5 & 6.68 &[0.260, 0.367, 0.600, 0.765] \\
        \hline
        edgar\_army & 6 & 4.69 & [0.260, 0.450, 0.636, 0.822]\\
        \hline
        edgar\_classroom & 6 & 5.27 &[0.300, 0.450, 0.687, 0.925]\\
        \hline
        outdoors & 9 & 3.87 &[0.474, 0.875, 1.374, 1.736]\\
        \hline
        \hline
        Overall & 44 & 4.39 & [0.335, 0.497, 0.862, 1.152] \\
        \hline
    \end{tabular}
        \caption{Comparison of lidar and radar maps on the ColoRadar dataset. The ratio of points in the lidar map to the radar map L/R is provided along with the percentage of points contained within a certain deviation percentage in meters. For example, in the ARPG Lab Scene, 95\% of radar points are within 1.5m of a lidar point.}
        \label{tab:dataset}
    \vspace{-7mm}
\end{table}

Considering the overall L/R ratio was approximately 4, we generated input patches with 2048 points and ground-truth patches with 8192 points using pose sampling. We consider the scene, labeled run0, of each environment as test data and the rest of the scenes, run1-run($x$), as train data. 

We trained with a set of hyperparameters similar to those used in the AdaPoinTr backbone. A notable exception is that AdaPoinTr added in 64 noise points to their queries induced in the Dynamic Query bank. We instead set the number of added points to 0. The resulting coarse point cloud generated from AdaPoinTr $\mathcal{P}_{c,0}$ has 512 points. As a result, we use the upsampling factors of [1, 4, 4] in the upsample layers to generate the final pointcloud with 8192 points.

We trained UpPoinTr end-to-end with AdamW\cite{loshchilov2018fixing} optimizer with an initial learning rate of $10^{-5}$ and weight decay of $5 \times 10^{-5}$. The batch size is set to 16, and trained for 600 epochs with a continuous learning rate decay of 0.9 for every 20 epochs.



\section{Results}
\label{sec:Results}
As a baseline for comparison between our original dataset and our predicted radar maps, we present the deviation distribution metric for the entire ColoRadar dataset in Table \ref{tab:dataset}. The metric provides the distribution of the distance from a point in the lidar point cloud to the nearest radar point in the radar point cloud, as defined in Equation \ref{eq:distribution}. Based on this metric $<50\%$ of the points in the initial radar point cloud are contained within the voxel resolution of 0.15m. At the $75$th percentile most points in the radar map are at least 3 voxel widths apart at $0.45$m. 

\subsection{Evaluation Metrics}
\begin{table*}[t]
    \vspace{5mm}
    \centering
    \begin{tabular}{|p{2cm}||p{1cm}|p{1cm}|p{4cm}||p{1cm}|p{1cm}|p{4cm}|}
        \hline
        \textbf{Test Environment} & \textbf{Input CD-$l1$} &  \textbf{Input CD-$l2$} & \textbf{Deviation distribution [50\%, 75\%, 90\%, 95\%] (m)} & \textbf{RMap CD-$l1$} &  \textbf{RMap CD-$l2$} & \textbf{Deviation distribution [50\%, 75\%, 90\%, 95\%] (m)} \\
        \hline \hline
        arpg\_lab & 0.329 & 0.651 & [0.212, 0.367, 0.618, 0.808] & 0.197 & 0.205 & [0.091, 0.127, 0.295, 0.499] \\ 
        \hline
        aspen & 0.500 & 1.184 & [0.335, 0.474, 0.687, 0.862] & 0.249 & 0.381 & [0.069, 0.123, 0.274, 0.454] \\
        \hline
        ec\_hallways & 0.426 & 0.717 & [0.367, 0.618, 0.900, 1.142] & 0.211 & 0.238 & [0.073, 0.115, 0.261, 0.528] \\
        \hline
        edgar\_army & 0.370 & 0.712 & [0.300, 0.450, 0.618, 0.765] & 0.175 & 0.155 & [0.067, 0.093, 0.146, 0.256] \\
        \hline
        edgar\_classroom & 0.388 & 0.786 & [0.300, 0.474, 0.704, 0.924] & 0.186 & 0.195 & [0.080, 0.103, 0.157, 0.249] \\
        \hline
        outdoors & 0.540 & 1.136 & [0.424, 0.765, 1.255, 1.650] & 0.357 & 0.644  & [0.093, 0.137, 0.381, 0.725] \\
        \hline
    \end{tabular}
        \caption{Comparison of input radar maps and the RMap or the maps generated from the outputs of UpPoinTr w.r.t lidar maps. Across all categories RMaps outperform the input radar map, improving the overall Chamfer Distance metrics and reducing the percentage of points in the radar map that occur outside 0.15m or one voxel of the nearest lidar point.}
        \label{tab:coloradar_results}
        \vspace{-5mm}
\end{table*}

We consider the CD-$l_1$, CD-$l_2$, and F-Score, commonly used in evaluating the point completion networks on ShapeNet and PCN dataset. Lower CD metrics and higher F-Scores are considered better. CD-$l_1$ uses an $l_1$ norm to calculate the distance between points in Chamfer distance, while CD-$l_2$ uses an $l_2$ norm. Both CD metrics are scaled by 1000, as is typical in other evaluations \cite{yuan2018pcn,yu2021pointr,yu2023adapointr,zhou2022seedformer}. We also calculated the F-Score following \cite{tatarchenko2019single}:

\begin{align}
     P(d) &= \frac{1}{|P|} \sum_{p \in P} [ \min_{g \in G} \| p - g \| < d ]
\end{align}
\begin{align}
    R(d) &= \frac{1}{|G_i|} \sum_{g \in G} [ \min_{p \in P} \| p - g \| < d ]
\end{align}
\begin{align}
    \text{F-Score}(d) = \frac{2 \, P(d) \, R(d)}{P(d)+R(d)}
\end{align}
where $P(d)$ denotes the Precision and $R(d)$ denotes the Recall of the predicted points within a distance of $d$ (=0.01) from the ground truth points. In this equation, $P$ denotes the final predicted point cloud and $G$ denotes the ground truth point cloud. 

We also introduce a new metric to analyze the deviation distribution over the entire scene, providing deeper insights on the final predicted map compared to the initial radar map.
\begin{align}
    D &= \{\min_{p \in P}\| p - g \| \}  \quad \forall {g \in G}
    \label{eq:distribution}
\end{align}
We present the distribution of this deviation D at 50, 75, 90, 95 percentiles. Lower deviation distributions are considered better. 

\subsection{Quantitative Results}
\begin{table}
    \centering
    \begin{tabular}{|p{3cm}||p{1.7cm}|p{1cm}|}
        \hline
        \textbf{} & \textbf{F-Score} & \textbf{CD-$l_2$}\\
        \hline \hline
        AdaPoinTr \cite{yu2023adapointr} & 0.414 & 1.469  \\ 
        \hline
        \textbf{UpPoinTr (Ours)} & \textbf{0.441} & \textbf{1.386}  \\ %
        \hline
    \end{tabular}
        \caption{Relative performance of AdaPoinTr compared with UpPoinTr on the 3D ColoRadar map patches }
        \label{tab:adapointr_results}
        \vspace{-5mm}
\end{table}


\begin{table}
    \centering
    \begin{tabular}{|p{3cm}||p{1.7cm}|p{1cm}|}
        \hline
        \textbf{} & \textbf{F-Score} & \textbf{CD-$l_1$}\\
        \hline \hline
        PCN \cite{yuan2018pcn} & 0.695 & 9.64 \\
        \hline
        PoinTr \cite{yu2021pointr} & 0.745 & 8.38\\
        \hline
        SeedFormer \cite{zhou2022seedformer} & - & 6.74 \\
        \hline
        AdaPoinTr \cite{yu2023adapointr} & 0.845 & 6.53  \\ 
        \hline
        AdaPoinTr (w/o Denoising) & 0.810 & 6.92  \\ 
        \hline
        \textbf{UpPoinTr (ours)}& \textbf{0.848} & \textbf{6.47}  \\ %
        \hline
    \end{tabular}
        \caption{Relative performance of different networks on PCN dataset. We note that we do not provide CD$_{l2}$ as these methods were compared using reported scores from each method.}
        \label{tab:pcn_results}
        \vspace{-7mm}
\end{table}

We present each of these metrics in Table \ref{tab:coloradar_results} for both the test set and the predicted output of our network on the test set. Through these metrics, there is a clear increase in the number of points falling within the chosen voxel resolution of 0.15 with most environments maintaining this level of overlap out to the 75th percentile. Additionally, extreme errors, e.g. radar voxels over three voxel widths away from the nearest lidar voxel, occur around the 95th percentile in most datasets. The CD metrics also demonstrate a decrease in the trained datasets. 

We also compared our training on the ColoRadar dataset with AdaPoinTr. The results of this comparison are presented in Table \ref{tab:adapointr_results}. Our modified network, UpPoinTr, outperformed AdaPoinTr in both the F-Score metric and the CD-$l_2$ metric. 

Lastly, we compared our network along with a variety of other networks on the PCN dataset to test UpPoinTr's ability to generalize to different tasks. The results of this test are shown in Table \ref{tab:pcn_results}. Here, UpPoinTr proves marginally better than the next best architecture AdaPoinTr in both F-Score and CD-$l_1$ as well. 


\subsection{Qualitative Results}

\begin{figure}[tb]
    \centering
    \includegraphics[width=0.48\textwidth]{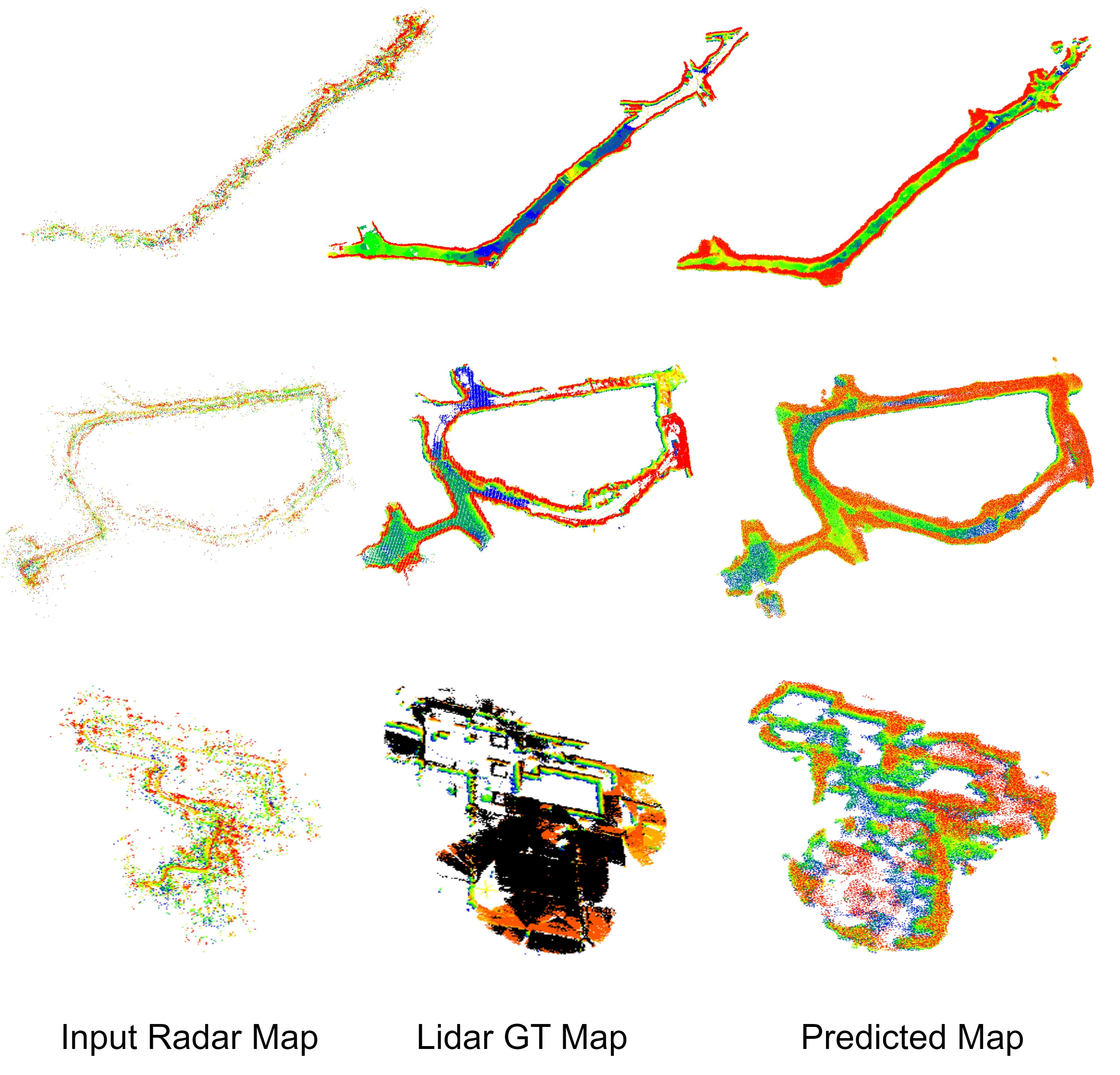}
    \caption{Comparison of generated maps with naive radar accumulation (1st column), lidar (2nd column), and our technique (3rd column). For all maps we have removed the ceiling at an arbitrary 0.8m height.}
    \label{fig:results}
    \vspace{-5mm}
\end{figure}

\begin{figure}[ht]
    \vspace{2mm}
    \centering
    \includegraphics[width=0.48\textwidth]{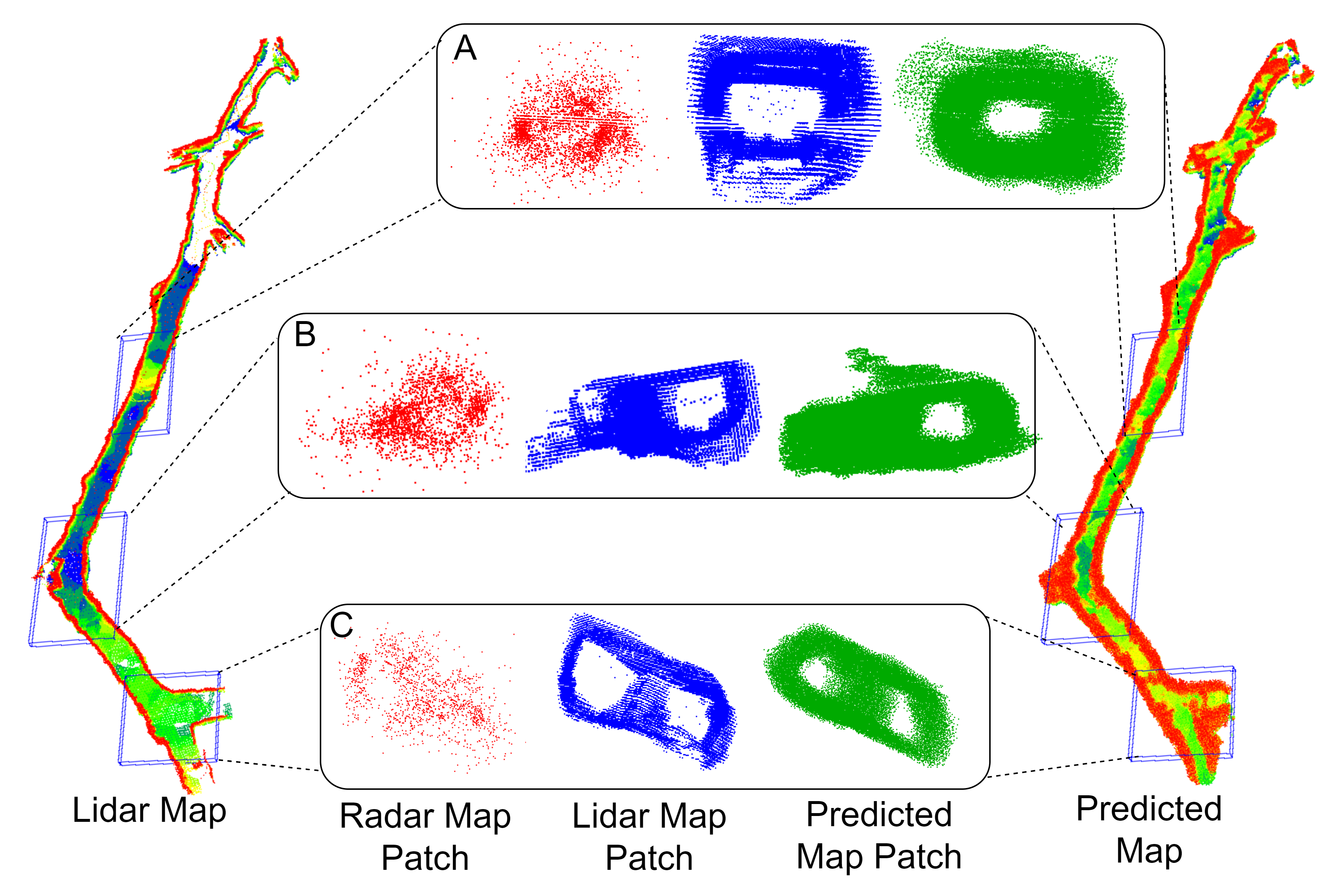}
    \caption{Detailed view of various cross sections on a tunnel (edgar\_army) map. (A) represents a cross-section of the maps along the straight path, (B) represents a cross-section of the maps at a corner, (C) represents a cross-section of the maps at a multi-path. Red point clouds represent the input radar map, which contains significant noise. Blue represents the lidar map detailing the actual structure of the environment with minimal noise. Green represents the predicted RMap. }
    \label{fig:crosssection}
    \vspace{-7mm}
\end{figure}

\begin{figure}[ht]
    \vspace{2mm}
    \centering
    \includegraphics[width=0.45\textwidth]{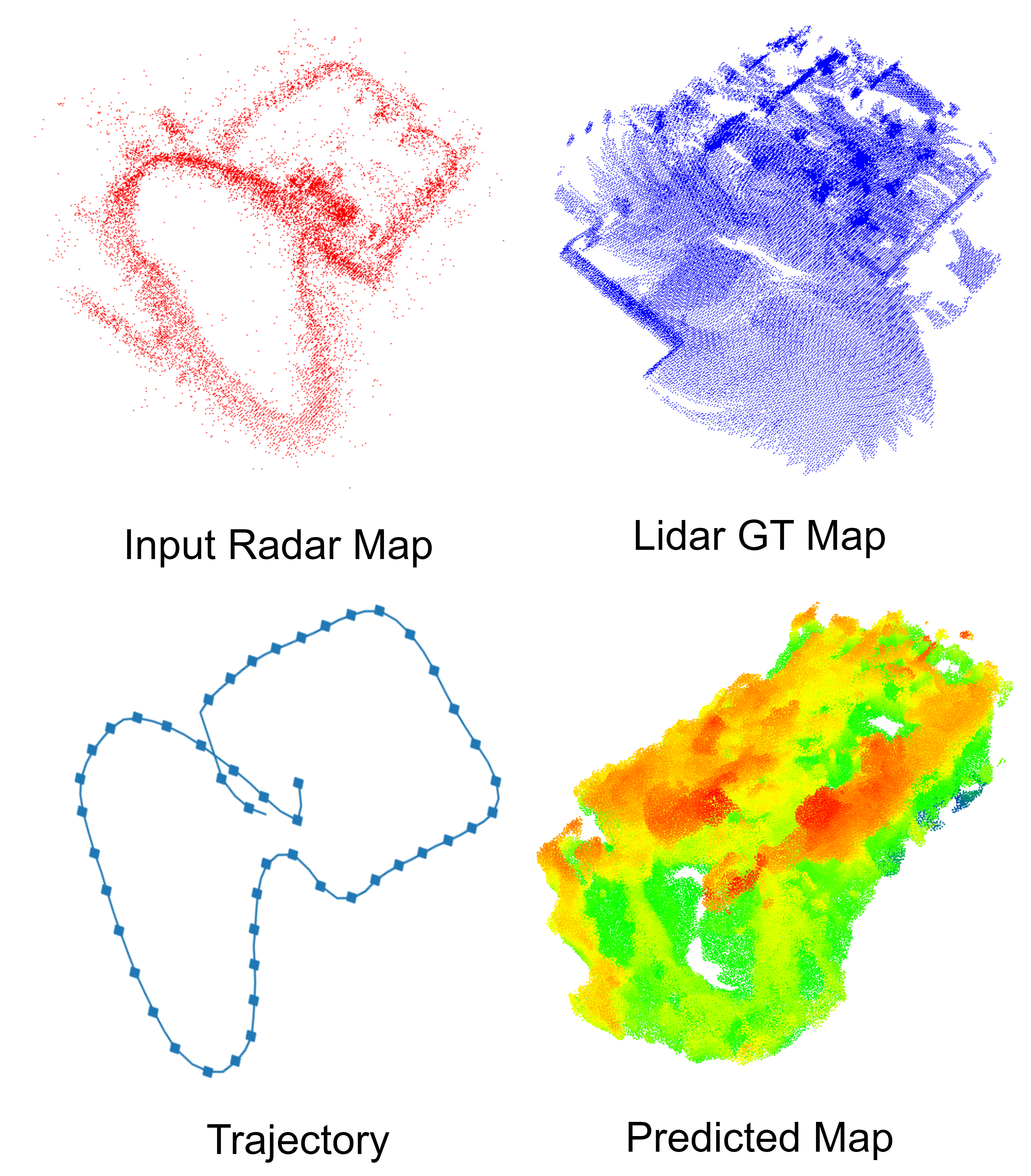}
    \caption{An example degenerate case from RMap. As our method acts as an infilling method in some regions, along the trajectory, in otherwise free space, RMap generates geometry along the ground (yellow) instead of maintaining a flat region. This can be seen in the elevation of the predicted map colored from low (green) to high (red). This is likely a result of only having radar points generated along the ground. In the structured portions of the environment, RMap still sufficiently reconstructs the map.}
    \label{fig:freespace}
    \vspace{-5mm}
\end{figure}

We show several example maps generated from merged predicted patches in Figure \ref{fig:results}. In general, the predicted maps successfully resemble the lidar maps -- tracking walls, free space, and clutter accurately.

More specifically we note in Figure \ref{fig:crosssection} the explicit benefits and some minor drawbacks of our method. This figure shows the cross-section of the original radar map, the lidar ground truth map, and the predicted radar map. Through this cross-section analysis, we see that the original radar map consists primarily of noise. However, after training it has a similar structure to the lidar map, distinguishing between free space and occupied space. 

We also note an example degenerate case in our method in Figure \ref{fig:freespace}. Our method acts as an infilling predictor. In this outdoor scene, the radar only detects points on the ground along the trajectory for a significant portion of the map. In the predicted map, these points along the floor are infilled with some additional geometry not present in the lidar groundtruth map. Conversely, the portion of trajectory in the structured area still performs well. We discuss some potential solutions to this in Section \ref{sec:discussion}.

\section{Discussion}
\label{sec:discussion}
We find RMap to be a compelling method for upsampling, filling, and denoising radar map points. In various quantitative results, our performance matches that of other point-cloud completion networks, which are primarily trained on synthetic pointcloud data rather than radar data. Notably, our network, UpPoinTr, even achieves state-of-the-art performance on general point cloud completion tasks as noted in Table \ref{tab:pcn_results}. 


The value of RMap becomes evident in the qualitative results, as depicted in Figure \ref{fig:results}. In structured environments, RMap represents the nearby structure of various test environments. A more detailed examination of a tunnel map, as presented in Figure \ref{fig:crosssection}, reveals that while the original map is not suitable for navigation but our noisy predicted map has the potential for navigability. Clear distinctions between free and occupied space are apparent, which were absent in the original radar map.

While RMap performs well in structured areas, it can encounter challenges in regions with little structure.  We note in Figure \ref{fig:freespace} that along empty sections of the trajectory, RMap generates geometry in the same shape as the original trajectory. In the more structured parts of the environment, however, the predicted radar map is still sufficiently structured. We acknowledge that some future work, pre-processing the radar maps prior to input into our network, or an alternate training strategy with more explicit representations for planar features such as large flat spaces may further enhance our method. 

\section{Conclusion}
\label{sec:conclusion}
We have presented a state-of-the-art method for generating 3D radar maps through our novel upsampling network UpPoinTr. We additionally designed a novel pose-based region sampling algorithm for maps. Our network has out-performed state-of-the-art point cloud completion networks in key metrics on radar data. As mentioned in Section \ref{sec:discussion}, the resulting RMaps produce radar maps that maintain similar structure to lidar maps, with the ability to operate in potentially visually degraded environments. Future work may further address separating empty regions and identifying key geometry either prior to input into the network or as a constraint within the network to avoid infilling improperly. However, RMap is demonstrated to capably separate free and occupied space as well as remove outliers using radar alone. 

\section{Acknowledgements}
This work was supported through the DARPA Subterranean Challenge cooperative agreement HR0011-18-2-0043 and the National Science Foundation \#1764092.

\printbibliography

\addtolength{\textheight}{-12cm}   







\end{document}